# A comparative study on face recognition techniques and neural network

Meftah Ur Rahman

Department of Computer Science

George Mason University

mrahma12@masonlive.gmu.edu

## 1. Abstract

*In modern times, face recognition has become one of the key aspects of computer vision. There are at least two reasons for this trend; the first is the commercial and law enforcement applications, and the second is the availability of feasible technologies after years of research. Due to the very nature of the problem, computer scientists, neuroscientists and psychologists all share a keen interest in this field. In plain words, it is a computer application for automatically identifying a person from a still image or video frame. One of the ways to accomplish this is by comparing selected features from the image and a facial database. There are hundreds if not thousand factors associated with this. In this paper some of the most common techniques available including applications of neural network in facial recognition are studied and compared with respect to their performance.*

## 2. Introduction

Human beings can distinguish a particular face from many depending on a number of factors. One of the main objective of computer vision is to create such a face recognition system that can emulate and eventually surpass this capability of humans. In recent years we can see that researches in face recognition techniques have gained significant momentum. Partly due to the fact that among the available biometric methods, this is the most unobtrusive. Though it is much easier to install face recognintion system in a large setting, the actual implementation is very challenging as it needs to account for all possible appearance variation caused by change in illumination, facial features, variations in pose, image resolution, sensor noise, viewing distance, occlusions, etc. Many face recognition algorithms have been developed and each has its own strengths [1, 2].

We do face recognition almost on a daily basis. Most of the time we look at a face and are able to recognize it instantaneously if we are already familiar with the face. This natural ability if possible imitated by machines can prove to be invaluable and may provide for very important in real life applications such as various access control, national and international security and defese etc. Presently available face detection methods mainly rely on two approaches. The first one is local face recognition system which uses facial features of a face e.g. nose, mouth, eyes etc. to associate the face with a person. The second approach or global face recognition system use the whole face to identify a person.

The above two approaches have been implemented one way or another by various algorithms. The recent development of artificial neural network and its possible applications in face recognition systems have attracted many reasearcher into this field. The intricacy of a face features originate from continuous changes in the facial features that take place over time. Regardless of these changes we are able to recognize a person very easily. Thus the idea of imitating this skill inherent in human beings by machines can be very rewarding. Though the idea of developing an intelligent and self-learning may require supply of sufficient information to the machine. Considering all the above mentioned points and their implications I would like to gain some experience with some of the most commonly available face recognition algorithms and also compare and contrast the use of neural network in this field.

## 3. Background

Throughout the past few decades there have been many face detection techniques proposed and implemented. Some of the common methods described by the researchers of the respective fields are :

PCA:

In high-dimensional data, this method is designed to model linear variation. Its goal is to find a set of mutually orthogonal basis functions that capture the directions of maximum variance in the data and for which the coefficients are pairwise decorrelated [3]. For linearly embedded manifolds, PCA is guaranteed to discover the dimensionality of the manifold and produces a compact representation. PCA was used to describe face images in terms of a set of basis functions, or "eigenfaces". Eigenfaces was introduced early [4] on as powerful use of principal components analysis (PCA) to solve problems in face recognition and detection. PCA is an unsupervised technique, so the method does not rely on class information. In our implementation of eigenfaces, we use the nearest neighbor (NN) approach to classify our test vectors using the Euclidean distance[2].

MPCA:

One extension of PCA is that of applying PCA to tensors or multilinear arrays which results in a method known as multilinear principal components analysis (MPCA) [5]. Since a face image is most naturally a multilinear array, meaning that there are two dimensions describing the location of each pixel in a face image, the idea is to determine a mulitlinear projection for the image, instead of forming a one- imensional (1D) vector from the face image and finding a linear projection for the vector. It is thought that the multilinear projection will better capture the correlation between neighborhood pixels that is otherwise lost in forming a 1D vector from the image [2].

LDA:

Fisherfaces is the direct use of (Fisher) linear discriminant analysis (LDA) to face recognition [6]. LDA searches for the projection axes on which the data points of different classes are far from each other while requiring data points of the same class to be close to each other. Unlike PCA which encodes information in an orthogonal linear space, LDA encodes discriminating information in a linearly separable space using bases that are not necessarily orthogonal. It is generally believed that algorithms based on LDA are superior to those based on PCA. However, other work such as [7] showed that, when the training data set is small, PCA can outperform LDA, and also that PCA is less sensitive to different training data sets.

ICA:

When applying PCA to a set of face images, we are finding a set of basis vectors using lower order statistics of the relationships between the pixels. Specifically, we maximize the variance between pixels to separate linear dependencies between pixels. ICA is a generalization of PCA in that it tries to identify high-order statistical relationships between pixels to form a better set of basis vectors. In [8], where the pixels are treated as random variables and the face images as outcomes. In a similar fashion to PCA and LDA, once the new basis vectors are found, the training and test data are projected into the subspace and a method such as NN is used for classification. The code for ICA was provided by the authors for use in face recognition research [8].

Neural Network:

To model our way of recognizing faces is imitated somewhat by employing neural network. This is accomplished with the aim of developing detection systems that incorporates artificial intelligence for the sake of coming up with a system that is intelligent. The use of neural networks for face recognition has been shown by [9] and [10]. In [11], we can see the suggestion of a semi-supervised learning method that uses support vector machines for face recognition. There have been many efforts in which in addition to the common techniques neural networks were implemented. For example in [12] a system was proposed that uses a combination of eigenfaces and neural network. In [13], first The dimensionality of face image is reduced by the Principal component analysis (PCA) and later the recognition is done by the Backpropagation Neural Network (BPNN).

The goal of this study is to gain experience in the above mentioned methods and also implement some of these so that some form of comparison can be done among these.

## 4. Overview

For the commonly available algorithms it is important to gain some theoretical knowledge before their implementation and their pros and cons. Based on the continuous reading of related scientific papers, some of the implementation might not be pragmatic in time permitted. Because there are a lot of issues associated with this. For example, to get an understandable working simulation the tools needed may not be available freely i. e. the tools might be offered by some vendors. In that case their implementations already done by researchers will be needed to take into account. Even in that case I'll try to get a through understanding of how it can be implemented in future, what are the things that are assumed or variables fixed for the implementation etc.

There are mainly two approaches face recognition algorithms. One way is general algorithmic (PCA, LDA, ICA etc.) and another one is AI centric (e.g. Supervised and unsupervised learning methods such as SVM, Neural Networks etc.). One way to gain a rough understanding of these two approaches would be to select any two algorithms of these and then run algorithms on some sample data. There are many databases freely available online for these purpose. After going through some of the available methods and tools it became apparent that some of these would be too time consuming to actually go through them in depth. MATLAB seemed to be a good choice in this respect. This is a fourth generation programming language and a numerical computing environment widely used by educational and research organization througout the whole world. Though is a proprietary software released by MathWorks, it has a fairly strong user groups all around the world. The algorithms generally proposed to use for face recognition are many times implemented by experienced researchers and users. Sometimes they share the actual implementation with other users. In order to reduce the implementation time, two of these were chosen to be implemented in this paper. As the implementation tool, the latest release of MATLAB R2011b is chosen. To implement the experiments, there are also two toolboxes that are necessary along with the main environment of matlab: image processing toolbox and neural network toolbox. The first toolbox is needed in order to implement the first part i.e. face recognition based on Eigenfaces and neural network toolbox is needed to test the neural network based implementation of face recognition technique.

## 5. Results

The experiment was started by implementing the Eigenfaces method under matlab. As described earlier, the latest release was installed under windows 7, along with image processing and neural network toolbox. 'Face Recognition System' is a demo code set provided by Luigi Rosa in Matlab central [14]. All parts of the code provided are written in Matlab language (M-files and M-functions) with no P-files (protected executables). The demo code is run on a small subset of AT&T's "The Database of Faces" (formerly "The ORL Database of Faces") and provided in a directetory along with the source codes [15].

As described in [14], the algorithm that actually uses this eigenfaces method also employs Karhunen-Loeve algorithms in order to improve efficiency. The system functions by projecting face images onto a feature space that spans the significant vairations among known face images. The significant features are known as "eigenfaces" because they are the eigenvectors (principal components) of the set of faces. Through the UI, face images were collected into sets: every set or class includes a number of images for each person, with some variations in expression and in the lighting. When a new input image is read and added to the training database, the number of class is required. Otherwise, a new input image can be processed and confronted with all classes present in database. The number of eigenvectors chosen is equal to the number of classes. Befor starting the image processing, first we need to select input image. This image can be successively added to database or if a database is alreadey present, matched with known faces.

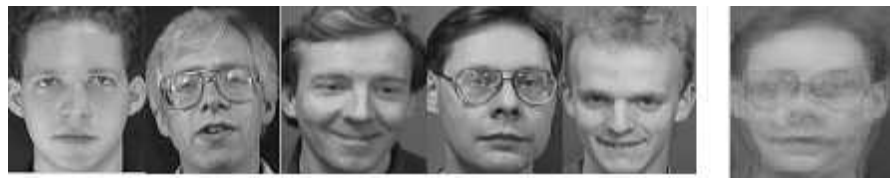

Figure 1: Samples of five classes along with their 'mean'

As shown in figure 1, sample images of five different images were added to the database one by one. After that the eigenface algorithm keeps mean of all these classes and continuously updates it as the databse is updated i. e. a new image is added over time.

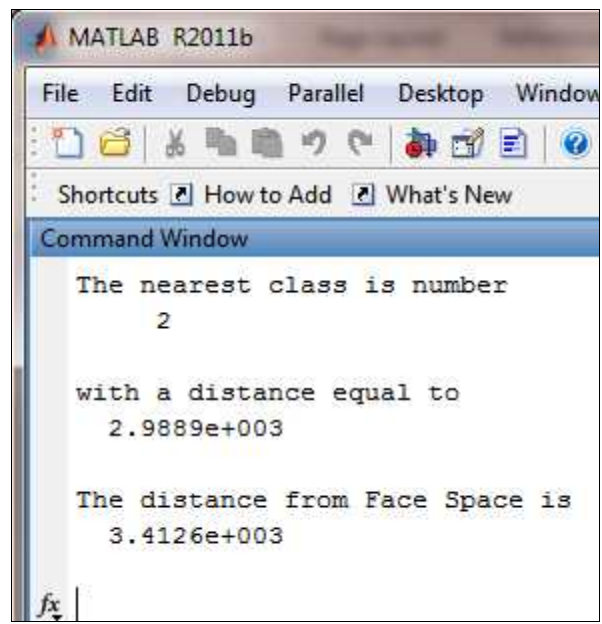

Figure 2 An instance of Matlab command window

When an image is selected and the face recognition function is working, the algorithm tries to calculate the distance of that particular image from the face space and returns the nearest class number to which the image might belong [fig. 2].

The second part of the experiment starts with an implementation of Face Detection using Gabor feature extraction and neural networks provided by Omid Sakhi on [16]. Before the actual implementation of neural network the set of sample image files need to go through gabor feature extraction. This program only detects faces tha tcan fit inside a 27x18 window. Initially it was made sure that neural networks and image processing toolbox both were installed with matlab.

First the network was trained and as soon as the network reached its predefined performance goal it stopped [fig.3].

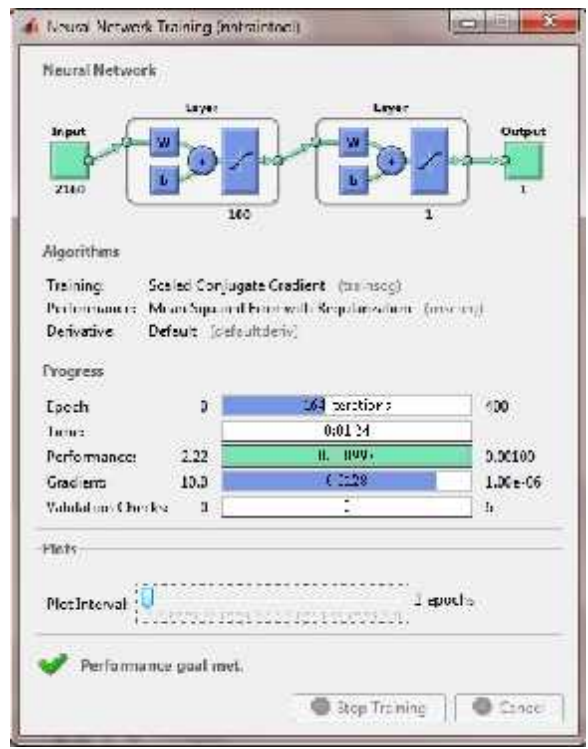

Figure 3 Training of neural network

Immediate after that, the backpropagation algorithm was implemented on various images. A set of sample images are provided in with the codes [fig.4].

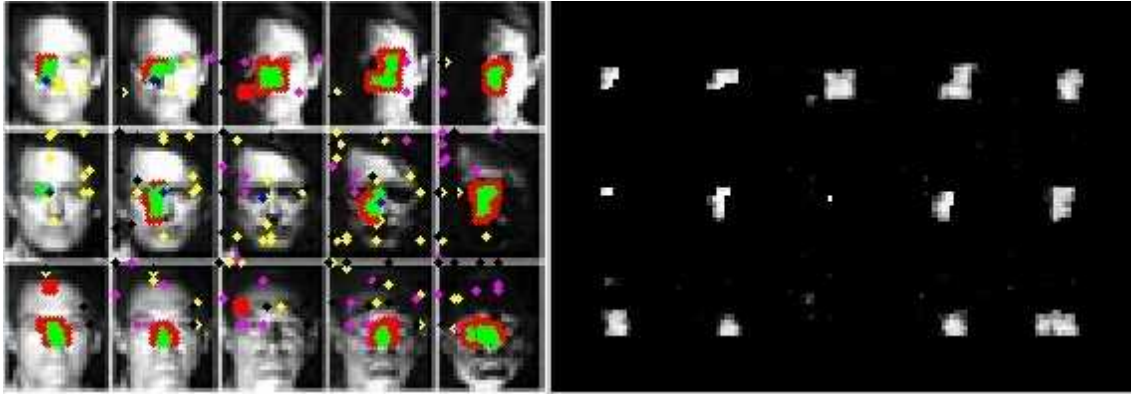

**Figure 4 Training of neural network**

Interpretation of Results:

For a number of input images (8/12 and 14/15 respectively), the correctness of eigenfaces method is approximately 66.67% and for neural network its 93.33%. For larger databases the correctness of eigenfaces may reduce somewhat. Because as the distance from the mean face for each individual becomes more densely distributed it becomes difficult for eigenfaces algorithm to distinguish in between so the result becomes more erroneous. Whereas in the case of neural networks, with more training and more complex neuron structure, its performace does not degrade so rapidly. In some cases we may see performance getting close to perfectness.

## 6.  Conclusion

The experiment has been done in a short period of time. Only two algorithms were analyzed in this paper. So from the result we can generalize in a rough scale. As many other issues were ignored to simplify the research scope, this generalization may not be entirely relevant to a real life dataset. Further research is possible to gain insights into the performance measurement and comparisons of other issues and algorithms as described in the earlier portions of the paper.

## 7.  References

[1] W. Zhao, R. Chellappa, A. Rosenfeld, P.J. Phillips; Face Recognition: A Literature Survey, ACM Computing Surveys, pp. 399-458, 2003

[2] Harguess, J., Aggarwal, J.K.; A case for the average-half-face in 2D and 3D for face recognition, IEEE Computer Society Conference on Computer Vision and Pattern Recognition Workshops, pp. 7-12, 2009

[3] Xiaofei He; Shuicheng Yan; Yuxiao Hu; Niyogi, P.; Hong-Jiang Zhang; , IEEE Transactions on Pattern Analysis and Machine Intelligence, pp. 328 – 340, 2005

[4] M. Turk and A. Pentland. Eigenfaces for recognition, Journal of Cognitive Neuroscience, 3(1), pp. 71–86, 1991

[5] H. Lu, K. N. Plataniotis, and A. N. Venetsanopoulos. Mpca: Multilinear principal component analysis of tensor objects. IEEE Trans. on Neural Networks, 19(1):18–39, 2008.

[6] P. N. Belhumeur, J. P. Hespanha, and D. J. Kriegman. Eigenfaces vs. fisherfaces: Recognition using class specific linear projection, In ECCV '96: Proceedings of the 4th European Conference on Computer Vision-Volume I, pages 45– 58, London, UK, Springer-Verlag. 1996.

[7] Martinez, A.M. ; Kak, A.C. ; IEEE Transactions on Pattern Analysis and Machine Intelligence, Volume : 23 , Issue:2, pp. 228 – 233, Feb 2001,

[8] M. S. Bartlett, J. R. Movellan, and T. J. Sejnowski; Face recognition by independent component analysis., IEEE Transactions on Neural Networks, 13:1450–1464, 2002

[9] Fan X. ; Verma, B. ; A comparative experimental analysis of separate and combined facial features for GA-ANN based technique, Sixth International Conference on Computational Intelligence and Multimedia Applications, 2005.

[10] Shaoning Pang; Daijin Kim; Sung Yang Bang; Face membership authentication using SVM classification tree generated by membership-based LLE data partition, IEEE Transactions on Neural Networks, Volume: 16 , Issue: 2 , pp. 436 – 446, 2005

[11] Ke Lu, Xiaofei He, Jidong Zhao; Semi-supervised Support Vector Learning for Face Recognition, Lecture Notes in Computer Science, pp. 104-109, 2006

[12] Jamil, N. ; Iqbal, S. ; Iqbal, N. ; Face recognition using neural networks, Technology for the 21st Century, pp. 277 – 281, IEEE INMIC, 2001

[13] P. Latha, Dr. L. Ganesan & Dr. S. Annadurai; Face Recognition using Neural Networks, Signal Processing: An International Journal (SPIJ) Volume (3) : Issue (5). 153, 2009

[14] http://www.mathworks.com/matlabcentral/fileexchange/4408

[15] AT&T Laboratories Cambridge. The ORL face database, Olivetti Research Laboratory available at http://www.uk.research.att.com/pub/data/att_faces.zip

[16] http://www.mathworks.com/matlabcentral/fileexchange/11073-face-detection-system

**Appendices**

i. MATLAB R2011b Documentation: http://www.mathworks.com/help/techdoc/index.html

ii. Image processing toolbar user guide: http://www.mathworks.com/help/toolbox/images/

iii. Neural Networks toolbox user guide: www.mathworks.com/help/pdf_doc/nnet/nnet_ug.pdf